\documentclass[11pt,a4paper]{article}
\usepackage[utf8]{inputenc}
\usepackage[T1]{fontenc}
\usepackage{mathptmx} 
\usepackage{microtype}
\usepackage[a4paper,margin=25mm]{geometry}
\usepackage{amsmath,amssymb,amsfonts}
\usepackage{graphicx}
\usepackage[affil-it]{authblk}
\usepackage{cite}
\usepackage[hidelinks]{hyperref}
\usepackage{orcidlink}

\title{\textbf{Facultative Swarming: A Capability, Not a Category, for Multi-Robot Systems}}

\author[1]{Ning Zhou\,\orcidlink{0009-0005-8352-1841}}
\author[1,*]{Edmund R. Hunt\,\orcidlink{0000-0002-9647-124X}}
\affil[1]{School of Engineering Mathematics \& Technology, University of Bristol, Ada Lovelace Building, Tankard's Close, Bristol BS8 1TW, United Kingdom}
\affil[*]{Corresponding author: \href{mailto:edmund.hunt@bristol.ac.uk}{edmund.hunt@bristol.ac.uk}}

\date{Preprint --- October 2026}

\begin{document}
\maketitle

\begin{abstract}
\noindent Collective robotics is split by a long-standing but increasingly artificial boundary between centralised multi-robot systems that prioritise optimality and decentralised swarms that prioritise robustness and adaptivity. We propose a unified control framework, shifting from `obligate' swarms---systems constrained to pure self-organisation---to facultatively swarming systems: hybrid architectures that operate as centrally planned systems under permissive conditions yet degrade gracefully into decentralised swarming under environmental stress. By reframing swarming as a dynamic capability, rather than a rigid system category, we outline a pathway towards collective robotic systems that benefit from centralised optimality when conditions permit and from the flexibility of self-organisation when they do not.

\end{abstract}

\section*{Introduction}

The deployment of multiple interacting robots has matured along two largely independent tracks. Multi-robot systems (MRS) leverage global communication and centralised planning to deliver efficient, verifiable coordination~\cite{parker2016multiple,gerkey2004formal}. Swarm robotics, by contrast, draws on biological metaphors---ant foraging, fish schooling, bird flocking---to achieve robust collective behaviour through strictly local sensing and peer-to-peer interactions~\cite{sahin2005swarm,Brambilla2013,reynolds1987flocks,Schranz2020}. Both traditions have produced landmark demonstrations. On the centralised side, warehouse fulfilment systems route hundreds of mobile robots through a central planner~\cite{wurman2008coordinating}, and motion-capture-guided fleets execute tightly choreographed aerial manoeuvres in dense formation~\cite{kushleyev2013towards,preiss2017crazyswarm}. On the decentralised side, swarms have achieved thousand-robot self-assembly~\cite{rubenstein2014programmable}, termite-inspired construction~\cite{werfel2014designing}, and heterogeneous platform coordination~\cite{dorigo2013swarmanoid}. Yet despite two decades of parallel progress, the robotics community has largely treated centralised MRS and decentralised swarms as distinct categories~\cite{dorigo2020reflections,yan2013survey,rizk2019cooperative}.

This dichotomy is increasingly untenable. Real-world deployment environments---warehouses with intermittent connectivity, disaster zones with degraded infrastructure~\cite{queralta2020collaborative,delmerico2019current}, contested theatres with adversarial jamming---do not respect the assumptions on which either paradigm was built. Centralised systems are brittle when communication channels saturate or fail~\cite{yan2013survey,rizk2019cooperative}; pure swarms sacrifice the global optimality and formal verifiability that industrial certification demands~\cite{lopes2016supervisory,dixon2012towards,Luckcuck2019}. A system designed for only one of these regimes is poorly matched to environments that oscillate between them.

Recent work has begun to erode this boundary. Research on adjustable and variable autonomy has established that a system's level of automation need not be fixed: control can be reallocated between human and machine, or across levels of autonomy, as task demands shift~\cite{parasuraman2000model,beer2014framework,dias2008sliding}. Yet, such approaches vary \emph{how much} autonomy an agent exercises, not \emph{how the collective coordinates}. The latter question is addressed directly by the self-organising nervous system (SoNS) architecture, which demonstrates that robots can autonomously establish, maintain, and reconfigure dynamic multilevel hierarchies, enabling temporarily centralised coordination within otherwise self-organised swarms~\cite{zhu2024sons}. Building on earlier work on mergeable nervous systems for physically connected robots~\cite{mathews2017mergeable}, SoNS allows agents to elect temporary `brain' robots that supervise sensing, actuation, and decision-making across a subswarm, while retaining the robustness, scalability, and flexibility characteristic of decentralised systems. This approach represents a major advance: it shows that hierarchy and self-organisation need not be antithetical. However, SoNS constructs its hierarchy from the bottom up---agents self-organise into a centralised structure. The question we pose here is complementary: what if the system begins centralised and must degrade gracefully into self-organisation when environmental conditions demand it?

\begin{figure}
\centering
\includegraphics[width=1\textwidth]{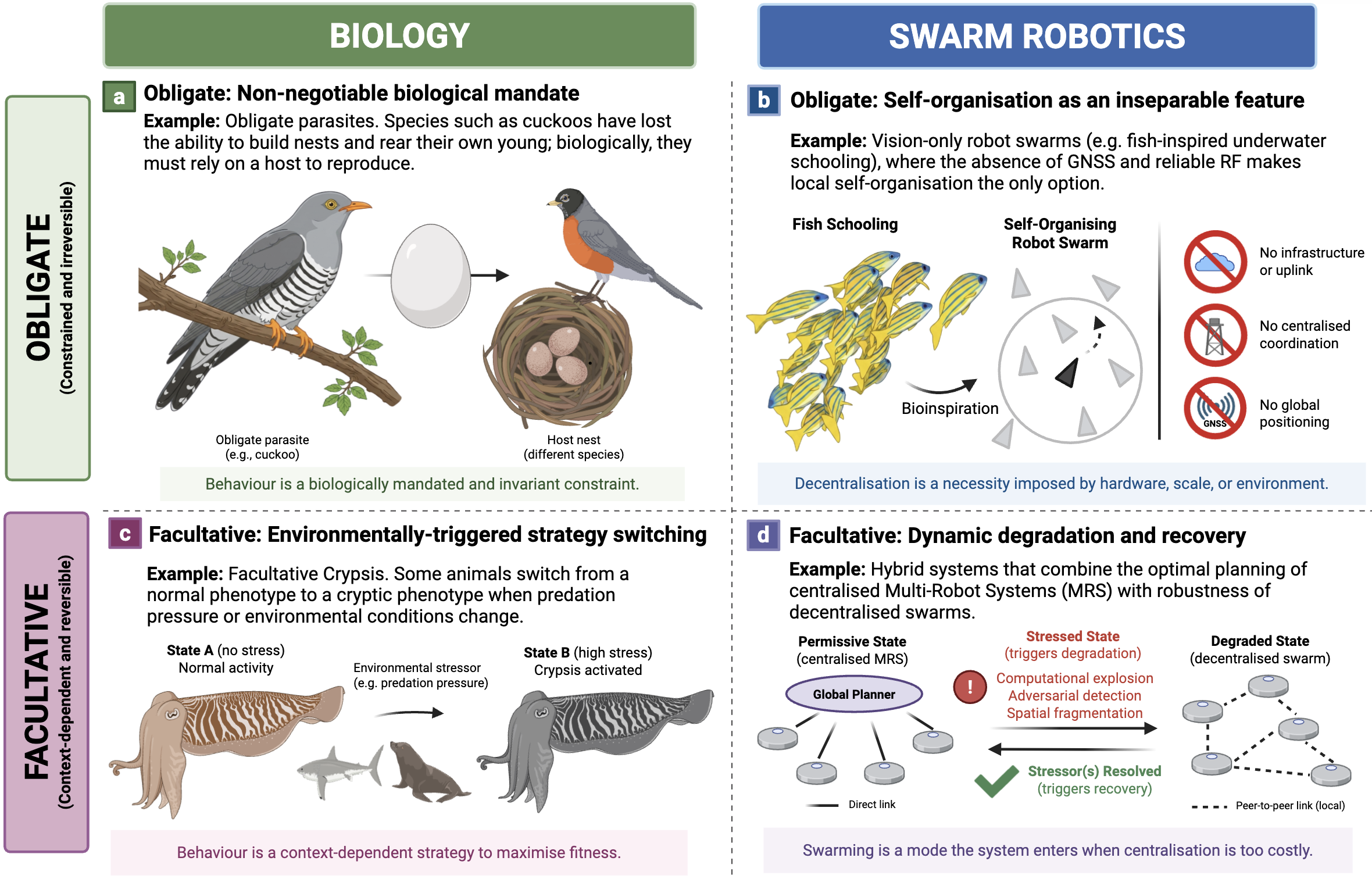}

\caption{
\textbf{Obligate versus facultative collective behaviour, across biology and swarm robotics.} The top row shows obligate behaviour (constrained and irreversible); the bottom row shows facultative behaviour (context-dependent and reversible). The left column gives biological examples; the right column, their swarm-robotics counterparts. The two biological cases are canonical exemplars of each mode rather than two ends of a single trait. \textbf{(a)}~The obligate brood parasite, which has lost the capacity for independent reproduction. \textbf{(b)}~Obligate robot swarms: self-organisation imposed by hardware, scale, or environment; here an underwater fish-inspired swarm~\cite{berlinger2021implicit}. \textbf{(c)}~Facultative crypsis: some animals can switch reversibly into a concealed state when a threat appears and abandon it when the threat passes---a context-dependent strategy rather than a fixed trait. Cuttlefish are a striking example, rapidly adjusting their appearance and behaviour to suppress the cues that would reveal them to a predator~\cite{bedore2015freezing}. \textbf{(d)}~A centrally planned MRS that degrades to decentralised self-organisation under stress, with example triggers across each of three categories, which recovers once the stressor(s) resolve. Created with BioRender.com.}

\label{fig:overview}
\end{figure}

To overcome this impasse, we borrow a conceptual lens from behavioural ecology. Ethologists classify traits by their necessity: an `obligate' behaviour is biologically mandated and invariant, whereas a `facultative' behaviour is a context-dependent strategy deployed to maximise fitness. The distinction is not a strict binary but a continuum: many behaviours grade from the opportunistic and conditional to the fixed and obligatory~\cite{piersma2003phenotypic}. Brood parasitism exemplifies this gradient~\cite{Sless2023}. A facultative brood parasite rears its own offspring but additionally lays eggs in others' nests when conditions favour it, retaining the full capacity for independent reproduction; an obligate brood parasite, at the far end of the continuum, has lost that capacity entirely and depends wholly on host nests to reproduce. We argue that applying this classification to robotic control architectures illuminates a productive way forward: the transition from obligate to facultative swarming (Figure~\ref{fig:overview}). 

\section*{The `Obligate' Constraint in Traditional Swarms}

The majority of physical robot swarms developed over the last two decades can be classified as obligate swarms. In these systems, decentralisation is not a strategic choice but a necessity imposed by hardware, scale, or environment. Minimalist research and education platforms such as the e-puck~\cite{mondada2009epuck} or r-one~\cite{mclurkin2013lowcost} lack the communication range and bandwidth for the global connectivity that centralised coordination assumes. Similarly, extreme scaling---the thousand-robot Kilobot swarm being the canonical example~\cite{rubenstein2014programmable}---renders centralised planning combinatorially intractable. In signal-denied environments---subterranean caverns, the deep ocean, or GNSS-denied urban canyons---the reliable, high-bandwidth communication that centralised control assumes is unavailable~\cite{Tranzatto2022,berlinger2021implicit,coppola2020surveymav}.

This obligate constraint has been scientifically productive. It has forced the community to develop effective local interaction rules, advanced our understanding of emergence, and generated a rich theoretical toolkit for analysing self-organised systems~\cite{camazine2001self,hamann2018swarm,Brambilla2013}. The findings can be counter-intuitive: constraining a swarm's communication can improve rather than degrade its collective adaptivity~\cite{talamali2021less}. However, the constraint also means that algorithms developed in obligate settings are rarely designed to exploit global information when it is available---which, in many deployment settings, is most of the time. An obligate swarm has no mechanism to capitalise on reliable connectivity when present, nor to exploit the transient opportunities---a passing relay, a temporary uplink, a brief clearing in RF interference---that arise even in nominally degraded conditions.

We argue that pure swarm robotics research---conducted under obligate constraints---remains essential. It is the design laboratory in which local coordination algorithms are conceived, refined, and formally characterised. However, for these algorithms to reach their full potential in real-world deployment, they must be integrated into broader architectures that can also exploit centralised planning when conditions permit. Self-organisation, in this view, is not a limitation to be overcome but a critical capability to be cultivated and then composed with complementary control strategies.

\section*{The `Facultative' Paradigm: Swarming as a Capability}

We propose \emph{facultative swarming} as a hybrid control capability that bridges the centralisation--decentralisation dichotomy. The paradigm presupposes platforms capable of centralised operation; it therefore complements rather than replaces the minimalist, hardware-obligate swarms that remain a distinct and valuable class \cite{mondada2009epuck,mclurkin2013lowcost}. A facultative swarmer retains the algorithmic and behavioural capacity to
function via local self-organisation when the cost of centralisation exceeds its benefit. Which regime is favoured at any moment is dictated not by the system alone but by its \emph{mission ecology}---the interaction of task demands, environmental conditions, and adversarial pressures that determines, moment to moment, what coordination the deployment can afford. This ecological framing has precedent in constraint-based robot ecology, where
survivability constraints are taken to shape robot behaviour over long deployments
\cite{Egerstedt2018,Egerstedt2021}; the facultative paradigm shares that intuition but
applies it to the collective's coordination regime rather than a single robot's control law.

In this paradigm, `swarming' is not what the system \emph{is}; it is a mode the system \emph{enters}. The architecture operates on a sliding scale of variable coordination regime, bounded by two canonical states. In the \emph{permissive state}, when bandwidth is abundant, latency is low, and the environment is predictable, the collective operates as a standard MRS: a global planner computes deterministic, near-optimal task allocations and trajectories~\cite{gerkey2004formal,khamis2015multirobot,guo2002distributed}. In the \emph{stressed state}, when triggered by specific environmental stressors, agents sever their reliance on the global planner and switch to local, implicit coordination protocols derived from the swarm-robotics literature~\cite{Brambilla2013,valentini2017bestn,mitrano2019minimalistic}.

The biological analogy is instructive. Facultative crypsis---the ability of certain prey animals to switch reversibly between conspicuous and cryptic phenotypes as conditions demand---illustrates the principle~\cite{polocavia2017pigmentation,bedore2015freezing}. An animal does not hide because it is broken; it hides because environmental context dictates that concealment is the optimal survival strategy, accepting costs such as higher energetic requirements in exchange~\cite{polocavia2017pigmentation}. Likewise, a facultatively swarming system does not decentralise because it has failed; it decentralises because the mission ecology has rendered centralisation suboptimal or unsafe.

Critically, the transition between modes need not be binary. Between the extremes of full central planning and full self-organisation lies a continuum of hybrid states. A cluster of agents encountering a local communication dead zone might switch to decentralised negotiation while the remainder of the collective continues under central control. Kloetzer and Belta observed that their centralised, verifiable swarm-control framework admits a decentralised variant in which each agent applies the same control law but computes the abstract state from its neighbours alone, recovering the centralised behaviour when neighbourhoods overlap sufficiently~\cite{kloetzer2007temporal}. Regional switching, hierarchical fallback, and graduated autonomy transitions are all architecturally feasible within the facultative framework, and each has partial analogues in existing multi-agent research: regional reconfiguration in market-based coalition formation~\cite{dias2006market}; hierarchical fallback in holonic control architectures, which reconfigure between hierarchical and heterarchical regimes~\cite{vanbrussel1998prosa}; and graduated transitions in the adjustable- and sliding-autonomy literature~\cite{beer2014framework,dias2008sliding}---there governing the human--machine allocation rather than, as here, the collective's coordination regime. The switching policy itself can be framed as a learning problem amenable to multiagent reinforcement learning \cite{zhu2024communication}.

Both the facultative paradigm and self-organising hierarchies such as SoNS~\cite{zhu2024sons} reject a shared assumption of conventional multi-robot systems: that the coordination architecture---its communication structure, control distribution, and degree of centralisation---is fixed before deployment, with robots merely coordinating within it. Where they differ is in the direction of reconfiguration. In SoNS, independent agents self-organise into a hierarchical structure by electing temporary leaders and granting them supervisory powers; the hierarchy emerges from the swarm. In a facultatively swarming system, the starting point is the opposite: a pre-existing centralised planner relinquishes control over specific agents or regions when environmental stressors render centralisation costly. SoNS answers the question `how can a swarm build hierarchy when it needs one?'; the facultative paradigm answers `how can a centrally planned system shed hierarchy when it cannot afford one?' These perspectives are not competing but convergent---they approach the same intermediate architectural space from opposite poles of the coordination regime. A mature collective robotic system might plausibly integrate both: using SoNS-like self-organised hierarchy to reconstitute coordination after a facultative degradation event, thereby enabling not only graceful degradation but also autonomous recovery.

\section*{Trigger Modalities: When to Swarm}
The core engineering challenge of the facultatively swarming system is the design of its meta-control switching layer: the logic that determines, in real time, when the cost or risk of maintaining centralised control outweighs the cost of decentralised organisation. Rather than enumerate a fixed list of triggers, we propose a taxonomy organised across three categories that together structure the space of conditions under which facultative switching is warranted. These categories are not mutually exclusive: a single stressor may implicate more than one.

\subsection*{A Taxonomy of Trigger Constraints}

\begin{figure}
\centering
\includegraphics[width=\textwidth]{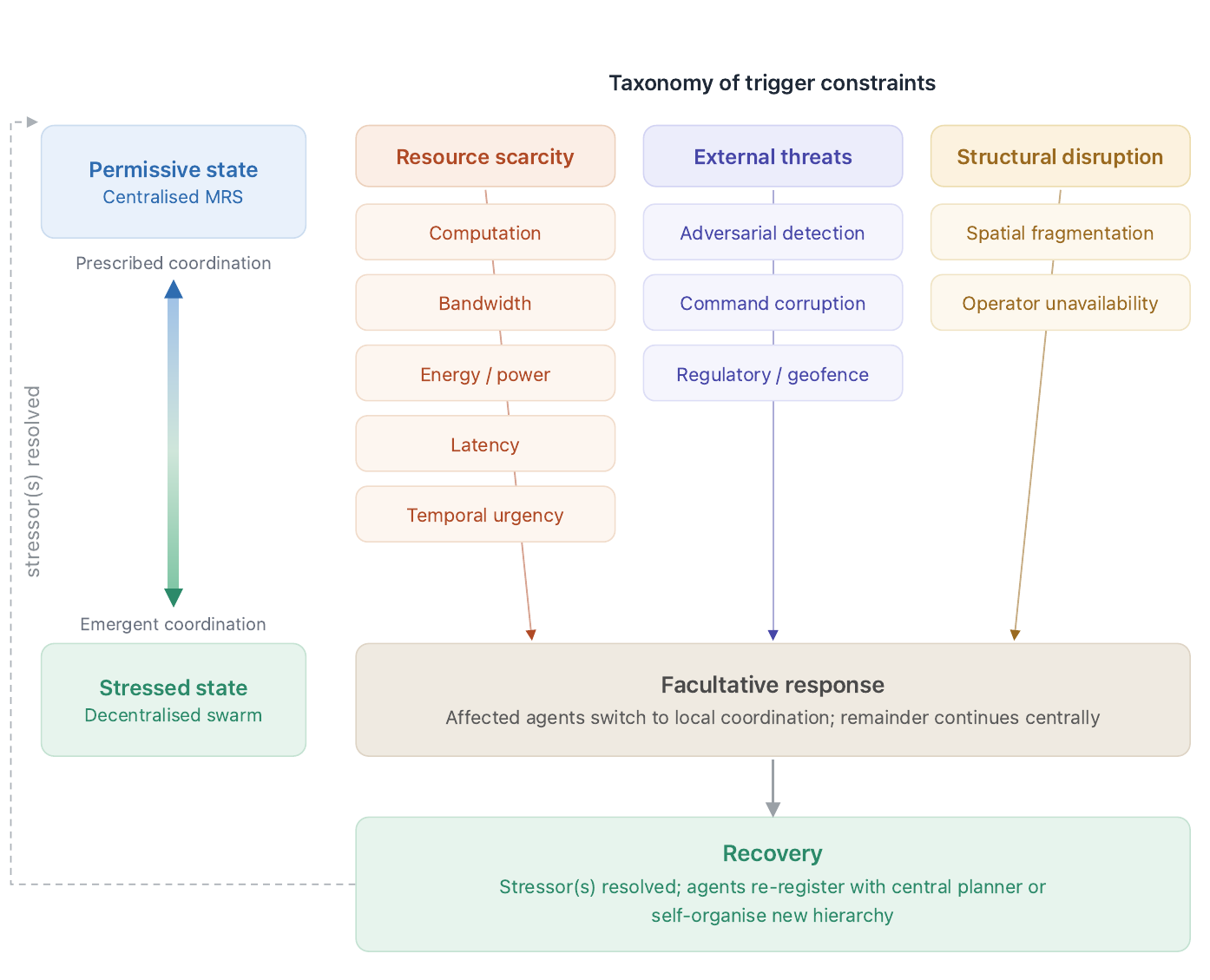}
\caption{\textbf{A taxonomy of trigger constraints for facultatively swarming.} The three categories across which centralisation can become untenable: \textbf{resource scarcity} (computation, bandwidth, energy, time), \textbf{external threats} (adversarial detection, command-channel corruption, regulatory prohibition), and \textbf{structural disruption} (spatial fragmentation, operator loss). Diagnosing which category (or categories) a stressor falls into identifies the relevant trigger modality and the switching logic a system designer must engineer. After switching to a decentralised swarm under stress, the system can recover centralisation when the stressors are resolved.}
\label{fig:taxonomy}
\end{figure}

The first category concerns \textbf{resource scarcity}: which resource required for centralised coordination has become limiting? Computation, bandwidth, energy, and time each represent distinct bottlenecks. The second category concerns \textbf{external threats}: is the centralised architecture itself a liability? Threats include adversarial detection of RF emissions, corruption of the command channel through cyber-attack, or regulatory prohibitions on certain communication modes. The third category concerns \textbf{structural disruption}: has the physical or organisational precondition for centralisation been lost? This includes spatial fragmentation of the collective and loss of the human operator who serves as de facto central planner. We develop a worked example within each category---two for resource scarcity, given the breadth of that category---before surveying the remaining modalities (Figure~\ref{fig:taxonomy}).

\subsection*{Resource Scarcity}

\subsubsection*{Computational Bottleneck}

In high-density commercial logistics, a central planner calculating collision-free paths for hundreds of autonomous mobile robots (AMRs) faces combinatorial explosion during localised disruptions~\cite{wurman2008coordinating,enright2011optimization}. A facultative architecture responds by switching agents in the congested sector to local, self-organised flow dynamics---goal-directed motion under local repulsion, in the spirit of social-force crowd models~\cite{helbing1995social}---while the remainder of the facility continues under central control. Once the disruption clears, affected agents re-register with the global planner. The trigger is computational: the planning cost for a local region has exceeded a defined threshold. We develop this trigger into a full proof of concept below.

\subsubsection*{Bandwidth Economy}

In data-intensive exploration missions, every coordination byte transmitted to the central planner competes with high-value payload data~\cite{guo2018multirobot,amigoni2017multirobot}. A facultative system can dynamically reallocate its RF spectrum: when payload acquisition intensifies, nearby agents switch to decentralised navigation (local potential fields, implicit repulsion), freeing bandwidth for high-resolution data transmission. The trigger is informational: the marginal value of a coordination byte has dropped below the marginal value of a payload byte.

\subsection*{External Threat}
\subsubsection*{Emission Control}
In contested environments, the radio-frequency signature of a centrally controlled MRS is a tactical vulnerability: the command-and-control link between agents and their operator can be passively detected, geolocated, and even classified by controller type from its emissions alone~\cite{Ezuma2020}. Agents can purposefully sever active radio links and transition to vision-based implicit coordination---relative-bearing tracking, leader-following via visual cues, or bio-inspired flocking~\cite{viragh2014flocking,vasarhelyi2018optimized,saulnier2017resilient}---drastically reducing their electronic footprint. The trigger is adversarial: active emissions have become a survivability risk. The move has biological counterparts: a cuttlefish facing an electroreceptive predator freezes to suppress the bioelectric field that would betray it~\cite{bedore2015freezing}, trading activity for stealth just as facultative swarming trades the command link for a reduced electronic signature.

\subsection*{Structural Disruption}
\subsubsection*{Spatial Fragmentation}
In missions that take a robot collective into buildings, underground, or behind terrain---urban search-and-rescue, subterranean exploration, contested-area reconnaissance---no single communication topology can be guaranteed to span the team for the duration of the mission. When a subgroup passes beyond line-of-sight or radio range of the central node, the precondition for centralised coordination is geometrically broken: there is no longer a connected path over which global state can be assembled or commands disseminated. This loss of connectivity is not a malfunction but an expected consequence of the mission---in subterranean exploration, robots are anticipated to range beyond the effective communication reach of the base, with winding passages and non-line-of-sight geometry severing the link~\cite{saboia2022achord}. A facultative architecture responds by letting the severed subgroup coordinate locally---maintaining cohesion and continuing its task through peer-to-peer interaction---until it re-enters communication range, directly or via a deployed relay, and reconciles its accumulated state with the central node. The trigger is geometric: the communication graph has fragmented, and no centralised world model can span the disconnected component.

\subsection*{Further Trigger Modalities}

The four triggers above are prominent examples, not a closed set. Within the resource-scarcity category, at least three further modalities merit attention. An \textbf{energy constraint} arises during battery-critical phases: agents in low-charge states drop the power-hungry uplink to the central planner and fall back on local coordination, conserving the transmission power that the central link would otherwise consume~\cite{Thammawichai2018}; the trigger is energetic rather than informational. A \textbf{latency constraint} arises in physically dispersed swarms (e.g., kilometre-scale aerial surveys) where round-trip communication delay to the central planner exceeds the local decision timescale demanded by obstacle avoidance or wind gusts; the channel has capacity, but the information is stale on arrival~\cite{hespanha2007survey}. A related \textbf{temporal-urgency constraint} arises when a sudden time-critical event---such as a detected casualty in search-and-rescue---creates a mismatch between the event timescale and the central planner's replan cycle; here it is not the channel but the planning loop that is too slow, and local agents must act immediately on local information.

Within the external-threat category, a \textbf{trust or integrity constraint} is triggered when the central node is suspected to be compromised---through cyber-attack, spoofed commands, or sensor injection. Unlike emission control, where the threat is detection, here the threat is corruption of the command channel itself; agents detect anomalous directives and facultatively reject centralised authority, falling back to resilient local consensus~\cite{leblanc2013resilient}. A \textbf{regulatory or geofence constraint} is triggered when agents enter a jurisdiction with legal restrictions on certain communication frequencies, autonomous coordination modes, or operational densities. The system could maintain centralised control but is prohibited from doing so; the trigger is normative rather than physical.

Finally, within the structural-disruption category, \textbf{human operator unavailability}---through fatigue, distraction, or ground-station link failure---triggers autonomous local coordination in human-in-the-loop systems where the operator serves as the de facto central planner~\cite{endsley1995toward,goodrich2008human,kolling2016human,Walker2012}.

This taxonomy is intended to be generative rather than exhaustive. By identifying the categories across which centralisation is failing---scarce resource, present threat, or lost structure---system designers can diagnose which trigger modalities are relevant to their domain and engineer the corresponding switching logic accordingly.

\subsection*{Formalising the Switch}
 
The decision to switch can be cast as a cost comparison, evaluated independently for each agent or spatial region~$i$, since the transition need not be collective: a congested sector may decentralise while the remainder of the collective continues under central control. Let $M_i(t) \in \{C, D\}$ denote the mode of region~$i$ at time $t$. The cost of centralised operation, $\mathcal{J}_C^i(t)$, is the nominal execution cost $E_C^i(t)$ plus a penalty for the stressors local to that region:
 
\begin{equation}
\mathcal{J}_C^i(t) = E_C^i(t) + \mathbf{w}^\top \mathbf{S}_i(t),
\label{eq:cost}
\end{equation}

where $\mathbf{S}_i(t)$ collects the resource, threat, and disruption triggers introduced above as sensed at~$i$, and $\mathbf{w}$ provides domain-specific weighting. The cost of decentralisation, $\mathcal{J}_D^i(t)$, combines the inherent sub-optimality of local rules, $E_D^i(t)$, with the anticipated overhead of eventual state reconciliation, $\gamma \Phi_{\mathrm{recon}}^i(t)$---the latter including the cost of whatever residual channel is needed to detect that recovery is warranted. 

What the cost comparison alone conceals is \emph{who evaluates it}. Applying a Schmitt-trigger formulation with asymmetric hysteresis thresholds ($h_{C \to D}$ and $h_{D \to C}$) to suppress chattering at the boundary $\mathcal{J}_C^i \approx \mathcal{J}_D^i$, the transition rule becomes

\begin{equation}
M_i(t+\Delta t) = 
\begin{cases} 
D & \text{if } M_i(t) = C \text{ and } \mathcal{J}_C^i(t) > \mathcal{J}_D^i(t) + h_{C \to D}, \\ 
C & \text{if } M_i(t) = D \text{ and } R_i(t), \\ 
M_i(t) & \text{otherwise,} 
\end{cases}
\label{eq:switch}
\end{equation}

where the degradation condition (first case) is computed by the central planner, which still holds global state while region~$i$ is centralised. The Schmitt-trigger rule is one natural instantiation; the switching policy could equally be learned or stochastic---for instance, posterior sampling over the two modes, or a soft switching rate, each recovering the deterministic threshold in its low-noise limit---without altering the structural conclusions below on where recovery authority resides. The recovery predicate $R_i(t)$ is satisfied when decentralisation has become the costlier option by the hysteresis margin, $\mathcal{J}_D^i(t) > \mathcal{J}_C^i(t) + h_{D \to C}$. Crucially, \emph{where} this predicate is evaluated depends on which stressor forced the switch. When the binding stressor lies within the resource-scarcity category, the central node persists and a thin monitoring channel survives, so $R_i$ can be evaluated centrally and the reconnect signal pushed to the region. When it lies within the structural or threat categories---spatial fragmentation, loss of the operator, a compromised command channel, emission risk---the centre is unreachable, untrusted, or re-contact is itself the hazard. In that case, $R_i$ cannot be centrally pushed and must instead be a locally computable proxy, ratified by local swarm consensus before the cluster re-forms, or used to seed a bottom-up reconstitution of hierarchy. For the threat case specifically, where the centre is untrusted rather than merely unreachable, the proxy must also re-establish trust before reconnecting—itself a non-trivial problem. 

This asymmetry is the formal counterpart of an architectural fact: degradation is always centrally orchestrated, because the planner is present by construction at the moment of shedding, whereas recovery authority migrates outward as the binding trigger moves from scarcity to threat to lost structure. The local recovery proxy accordingly never requires an agent to simulate full centralised cost; it is a threshold on a single locally observed signal---channel occupancy, a threat detector, a connectivity check---standing in for the active component of $\mathbf{S}_i$. A centrally signalled recovery does not eliminate the multi-agent transient but merely relocates it to the reconnect signal's propagation. The hysteresis bands and a minimum dwell time stabilise switching at the level of a single agent or sub-swarm; the proof of concept below operationalises the resource-scarcity case, and the residual multi-agent transient is taken up in the Challenges section.

\section*{A Proof of Concept: Facultative Planning under Computational Load}

\begin{figure}
\centering
\includegraphics[width=0.92\textwidth]{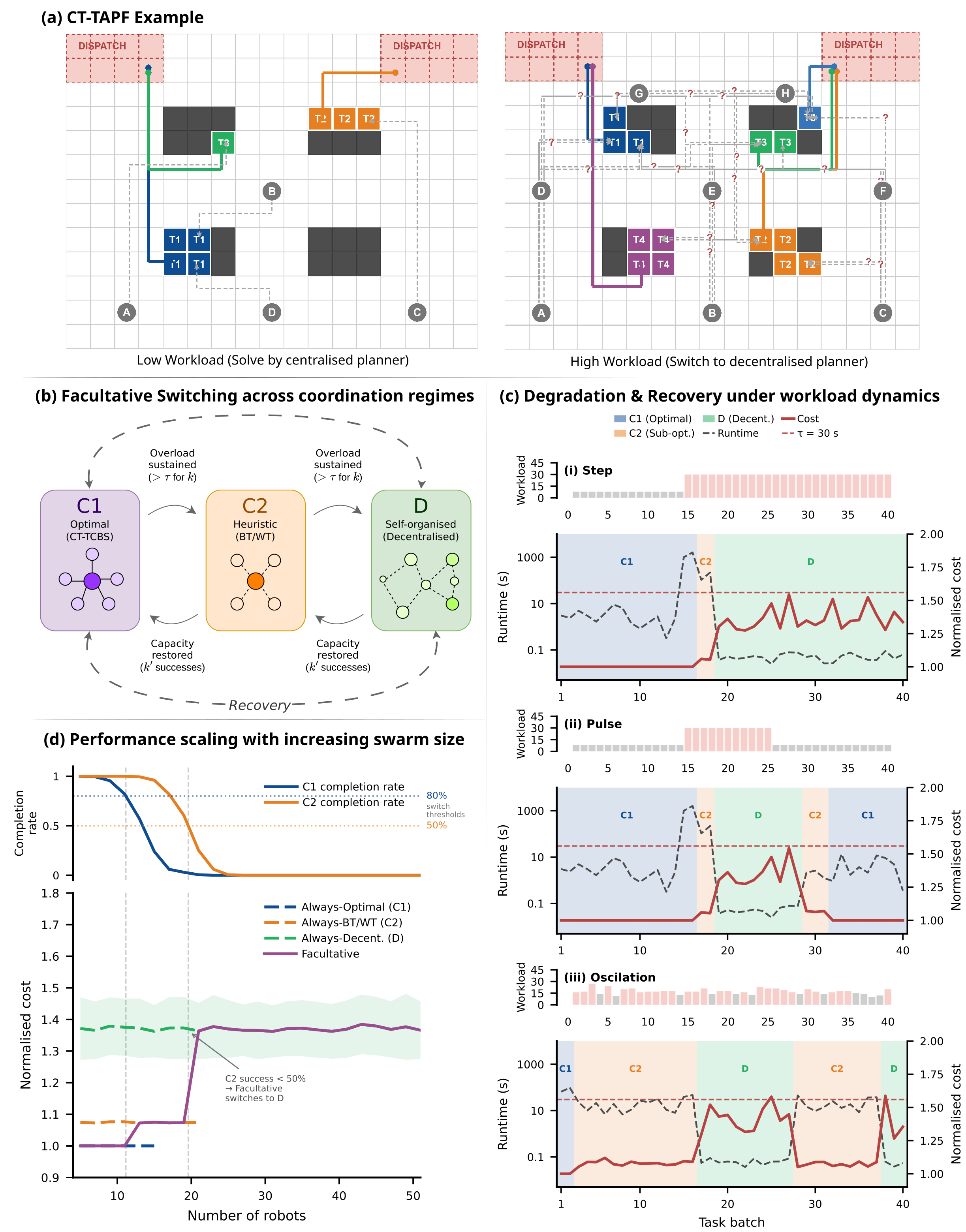}
\caption{\textbf{Facultative planning under computational load.} (a) The CT-TAPF testbed in the permissive (centralised, left) and stressed (decentralised, right) regimes. (b) The three-tier controller and its switching conditions. (c) Planner runtime (dashed) and normalised cost (solid) under step, pulse, and oscillating workloads, with the active mode shaded; representative runs of a simulation in which each mode's runtime and cost are sampled from distributions calibrated to CT-TCBS and GreedyPP benchmark runs~\cite{zhou2026cttapf}. (d) Completion rate and normalised cost versus team size from the same simulation; the facultative curve applies the switching rule to the per-mode completion rates, running the lowest-travel-cost mode that is still computationally tractable and stepping from C1 to C2 to D as each lower-cost mode in turn times out at scale.}
\label{fig:result}
\end{figure}

The computational-bottleneck trigger is not confined to physical motion planning; it recurs at the level of the planning algorithm itself, and it is here that we instantiate it. Our recent work on cooperative transportation task allocation and path finding (CT-TAPF) provides the testbed~\cite{zhou2026cttapf}: teams of agents are assigned to transport tasks---some requiring several robots to move a single large item---and routed along collision-free paths (Figure~\ref{fig:result}a). Within a centralised planner, optimality can itself be shed for tractability. An optimal centralised solver computes provably optimal team formations and trajectories, but the combinatorial cost of centralised team allocation grows explosively with problem size---a phenomenon we term the `task-conflict expansion dilemma'---and can be relieved by retreating to sub-optimal, task-centric heuristics that adopt a more local, greedy view~\cite{zhou2026cttapf}. This is already the facultative logic, applied to the planner rather than to the coordination locus; our proof of concept comprises both, degrading from an optimal central planner, through a heuristic central planner, to a fully decentralised mode.

Concretely, the collective runs in one of three coordination modes (Figure~\ref{fig:result}b). C1 is an optimal centralised planner (CT-TCBS) that returns provably optimal team assignments and trajectories. C2 is a sub-optimal heuristic planner (Best-Task/Worst-Task) that selects tasks from a global but greedy view~\cite{zhou2026cttapf}. D is a self-organised decentralised mode, in which agents abandon the global planner and coordinate through local rules.

The controller instantiates the cost comparison of Equation~\ref{eq:switch}, using planner runtime as the observable proxy for $\mathcal{J}_C$: a degrading transition fires only when overload is sustained, runtime exceeding $\tau$ for $k$ consecutive task batches, while a recovery transition fires once capacity is restored over $k'$ successful batches. The dwell time and hysteresis suppress chattering at the boundary.

Two simulation experiments characterise the policy. In both, each mode's per-batch runtime and solution cost are drawn from distributions calibrated to benchmark runs of the corresponding CT-TAPF solver~\cite{zhou2026cttapf}---the optimal CT-TCBS solver for C1, its Worst-Task heuristic for C2, and a greedy priority planner for D---and the switching logic is exercised on these draws; the traces shown are representative runs. The first experiment varies workload over time (Figure~\ref{fig:result}c). Under a sustained step in demand the controller degrades through C1, C2, and D and holds there; under pulse and oscillating demand it degrades and then recovers in step with load. In none does it chatter at the boundaries. The second varies the number of robots (Figure~\ref{fig:result}d). Here a mode's completion rate---the fraction of batches it clears within the runtime budget---is the scaling counterpart of the per-batch runtime trigger: as the collective grows, the completion rate of C1 falls first, then that of C2, crossing the $80\%$ and $50\%$ switch thresholds in turn, the asymmetry reflecting a greater tolerance for degradation before the more disruptive commitment to full decentralisation. At each team size the facultative policy runs the lowest-travel-cost mode that is still computationally tractable: it holds with C1 (optimal) while its planner keeps pace, falls back to C2 as optimal allocation begins to time out, and commits to D once C2 clears fewer than half its batches---each switch landing cleanly at the corresponding threshold. Each fixed strategy is in turn dominated, or fails outright, outside its own regime.

This controller resembles Dynamic Algorithm Configuration (DAC), which adjusts an algorithm's parameters online as a solve progresses~\cite{biedenkapp2020dac}. The computational-bottleneck trigger is the disembodied limit of the facultative paradigm; embodiment, however, is what separates the two. In DAC, a configuration change is free, instantaneous, and reversible, and acts on a single solver. In a swarm, the switch reconfigures physical agents that keep moving, opening a transient window in which neither mode's guarantees hold. Reverting is not free, because agents must reconcile drifted estimates and divergent task beliefs. And the triggers are exogenous and safety-bearing, such as jamming or emission risk, rather than internal measures of problem difficulty. The facultative paradigm thus generalises DAC from tuning one solver to reconfiguring a population of embodied agents. Each of these differences becomes an engineering challenge, which we turn to next.

\section*{Challenges and Open Questions}

The facultative swarming paradigm, while conceptually appealing, surfaces several non-trivial engineering and theoretical challenges that must be addressed for practical realisation.

\textbf{Commensurating the triggers.} Equation~\ref{eq:switch} presupposes that the resource, threat, and disruption stressors collected in $\mathbf{S}_i(t)$ can be weighted onto a single cost scale, yet a computational overrun, an emission-detection risk, and a regulatory prohibition are not naturally measured in common units. Calibrating the weight vector $\mathbf{w}$---deciding how to trade a quantifiable runtime cost against a probabilistic survivability risk---is itself an open problem, and one that governs whether the switching layer behaves sensibly across categories rather than only within the single category exercised by our proof of concept.

\textbf{Switching dynamics and stability.} The hysteresis rule of Equation~\ref{eq:switch} stabilises switching for a single agent or sub-swarm, but it does not resolve the multi-agent transient. When a switch propagates unevenly---some agents have changed mode while others have not---the collective enters a `window of vulnerability' in which neither centralised guarantees nor emergent coherence hold. Characterising and bounding this transient is an open problem.

\textbf{State reconciliation.} After a period of decentralised operation, agents must reconcile their local state estimates with the global planner upon re-centralisation. Drift in odometric estimates and stale map data must be efficiently fused~\cite{thrun2005probabilistic,Lajoie2024}, while divergent task beliefs must be reconciled to a conflict-free assignment~\cite{choi2009consensus}. This drift is not static but accumulates with time in the decentralised mode: reactive local control laws, in which each agent responds to its neighbours, incur bounded but non-zero error that grows with manoeuvring and swarm size. The reconciliation cost therefore scales with the duration of decentralised operation, and must be factored into the switching decision itself: premature re-centralisation may impose a reconciliation overhead that exceeds the benefit of returning to global planning, while prolonged decentralisation raises that overhead the longer it persists.

\textbf{Formal verification.} A key promise of centralised MRS is verifiability---the ability to certify system behaviour against a formal specification, for instance by mapping temporal-logic task specifications to provably correct control laws~\cite{kloetzer2007temporal}. Decentralised swarm behaviour, grounded in emergence, resists such certification~\cite{dixon2012towards,Luckcuck2019,winfield2005formal}. A facultative system must therefore provide safety guarantees that span both modes and, crucially, the transitions between them. Compositional verification frameworks that can reason about mode-switching hybrid systems are a pressing research need.
 
\textbf{Human--swarm interaction.} If the trigger for mode switching is partly or wholly delegated to a human operator, the cognitive demands of supervising a system that changes its own coordination regime are substantial~\cite{endsley1995toward,goodrich2008human,kolling2016human}. Transparent interfaces that communicate the current mode, the rationale for switching, and the expected behavioural consequences are essential to maintain operator trust and situational awareness.

\textbf{Behavioural heterogeneity on degradation.}
Under central control the planner supplies behavioural differentiation externally, so platforms can remain quasi-homogeneous; when control is shed, that heterogeneity could vanish precisely when the (sub-)swarm most needs intrinsic behavioural variety to handle conditions it can no longer centrally assess. Natural collectives anticipate their ecology by carrying a pre-tuned distribution of behavioural types (e.g.\ \cite{BurnsDyer2008}). A facultative system can be engineered analogously, by carrying a \emph{latent behavioural distribution} across the robots that is masked while the planner dictates behaviour and expressed only once control is relinquished. This distribution has two components that persist concurrently: an \emph{intrinsic baseline}, fixed before the mission and held in each robot's memory (e.g.\ \cite{Kengyel2015,York2024}), durable but contextually na\"ive; and an \emph{ad hoc planner update} layered over it while communication holds, conditioned on the evolving mission ecology and on a region's actual membership, geometry, and task assignment---contextually apt but frozen at the switch. On an unplanned disconnection the two differ in validity over time: in the short term the planner update best suits the robots' circumstances and they do well to retain it, but this snapshot decays as agents move, mix, and the local context drifts, so the robots should relax toward their intrinsic baseline. Specifying this baseline-and-update handover, and guaranteeing its coherence across the switching transient, remains open.

\section*{Conclusion}

The transition from obligate to facultative swarming represents an expansion of collective robotics---not a departure from swarm robotics but a broadening of its reach. The facultative paradigm builds directly on the local coordination algorithms that pure swarm research has spent decades refining, embedding them within a broader architecture that can also leverage centralised planning when conditions permit. Continued fundamental research in self-organised collective behaviour is therefore not merely compatible with the facultative vision but essential to it: every new swarm algorithm developed under obligate constraints enriches the behavioural repertoire available to a facultative system in its stressed state. The three trigger categories outlined here---resource scarcity, external threats, structural disruption---are not exhaustive; they are illustrative of the broader insight that swarming is most powerful when it is deployed as a context-dependent capability within a system that can also operate centrally. We have shown in simulation, for the computational-bottleneck trigger, that such switching can be made stable and chatter-free, but the remaining modalities await comparable demonstration. Realising the facultative swarming vision will require advances in key challenges such as switching dynamics, state reconciliation, and the anticipatory control of behavioural heterogeneity. The reward, however, is substantial: a class of collective robotic systems that draw on the optimality of centralised planning when conditions permit, and on the robustness, scalability, and flexibility of self-organisation when they do not---accessing the strengths of both regimes rather than being confined to one.

\section*{Funding}
E.R.H.\ is supported by the Royal Academy of Engineering under the Research Fellowship programme. N.Z.\ is supported by the China Scholarship Council.

\section*{Author contributions}
E.R.H.\ conceived the original concept, developed the theoretical framework, supervised the research, and wrote the first draft of the manuscript. N.Z.\ designed and implemented the simulation experiments, prepared the figures, and contributed to revising the manuscript.

\section*{Data availability}
The CT-TAPF solver and benchmark instances used to calibrate the simulation are those of~\cite{zhou2026cttapf}. The simulation scripts and data supporting the results reported in this manuscript are available from the corresponding author on reasonable request, and will be deposited in a public repository upon publication.

\section*{Competing interests}
The authors declare that they have no known competing financial interests or personal relationships that could have appeared to influence the work reported in this paper.

\bibliographystyle{ieeetr}
\bibliography{references_FS}

\end{document}